\documentclass[runningheads]{llncs}

\usepackage{eccv}

\usepackage{eccvabbrv}

\usepackage{graphicx}
\usepackage{booktabs}
\usepackage[table]{xcolor}
\usepackage{array}
\usepackage{multirow}
\usepackage{makecell}
\usepackage{arydshln}
\usepackage{xcolor}
\usepackage{pgf}
\usepackage{acro}

\usepackage[accsupp]{axessibility}  

\usepackage{hyperref}

\usepackage{orcidlink}

\begin{document}

\title{Understanding Cross-Rig Generalization in Automotive Perception: a Multi-Rig Benchmark and Rig Variation Metrics} 

\titlerunning{Understanding Cross-Rig Generalization in Automotive Perception}

\author{Tim Alexander Bader\inst{1,2}\orcidlink{0009-0006-3391-277X} \and
Tim Dieter Eberhardt\inst{1,2}\orcidlink{0009-0001-7079-9220}\and \\ 
Maximilian Dillitzer\inst{1,3} \and
Wilhelm Stork\inst{2}\orcidlink{0000-0003-0579-4615}}

\authorrunning{Bader et al.}

\institute{Dept. of Highly Automated and Assisted Driving, Dr. Ing. h.c. F. Porsche AG, Porscheplatz 1, Stuttgart, 70435, Baden-Württemberg, Germany\\
\email{\{tim.bader1,tim.eberhardt1,maximilian.dillitzer\}@porsche.de}\and
Institute for Information Processing Technologies, Karlsruhe Institute of Technology, Kaiserstraße 12, Karlsruhe, 76131, Baden-Württemberg, Germany\\
 \and
Faculty of Mobility and Technology, Esslingen University of Applied Sciences, Kanalstraße 33, 73728 Esslingen am Neckar Esslingen, Germany}

\maketitle

\begin{abstract}
Camera-based perception systems for autonomous driving are typically developed and evaluated using fixed sensor rigs, while real-world vehicle fleets exhibit substantial variation in camera placement, orientation, field of view, and camera count. 
This mismatch introduces a cross-rig domain gap in which only the geometric observation process changes. 
To study this effect under controlled conditions, we introduce \textit{Plentiful CARLA Camera Rigs}, a benchmark that renders identical driving scenes under 14 systematically designed camera rigs. 
This setup enables direct analysis of cross-rig generalization without confounding changes in scene content or appearance. 
Using the benchmark, we analyze cross-rig transfer behavior of representative multi-view perception architectures and observe substantial performance shifts induced by geometric rig variation. 
To facilitate structured analysis, we further introduce two calibration-based descriptors derived from rig metadata: \textit{Rig Variance}, capturing internal rig diversity, and \textit{Rig Contrastive Distance}, measuring geometric discrepancy between rigs. 
Our experiments show that geometric rig differences strongly correlate with relative cross-rig performance shifts and that Rig Contrastive Distance provides a reliable proxy for ranking transfer difficulty between sensor rigs.
\end{abstract}    
\section{Introduction}
\label{sec:intro}

Camera-based perception systems for autonomous driving have achieved strong performance under carefully engineered and largely fixed sensor configurations.
In other words, during both training and evaluation, a specific number of cameras with fixed positions, orientations, and fields of view is required~\cite{caesar2020nuscenes, sun2020scalability, nvidia_physicalai_av_2025}.
In practice, however, vehicle fleets are heterogeneous.
Differences in vehicle segment, packaging constraints, cost targets, and generational updates lead to variations in camera placement, overlap, field of view (FOV), and even camera count.

This setting raises a fundamental question:
\emph{how do camera rig characteristics alone influence perception performance?}
While this problem is often implicitly absorbed into broader notions of domain shift, cross-rig transfer differs in an important way: only the geometric observation process varies.
Performance shifts therefore arise purely from differences in viewpoint distribution, coverage, and redundancy.
We refer to this phenomenon as the \emph{cross-rig domain gap}.

The idea of decoupling perception models from specific sensor rigs has received only limited attention in prior work~\cite{reichert_towards_2021, bader2026universal}. 
What exists is Visual Question Answering (VQA) across two datasets with different sensor rigs~\cite{sima2024drivelm} and the observation of a \emph{cross-sensor domain gap}~\cite{klinghoffer_bevviewpointrobustness_2023, embacher_neural_2025}. 
However, most benchmarks typically assume fixed sensor configurations~\cite{liu2024datasetsurvey, bader2026universal} or limited viewpoint perturbations~\cite{embacher_neural_2025, klinghoffer_bevviewpointrobustness_2023}, and cross-dataset evaluations conflate rig changes with scene statistics, appearance differences, and dataset bias. 
As a result, there is currently no comprehensive protocol to disentangle performance variation caused specifically by rig characteristics.

\begin{figure}[t]
  \centering
  \includegraphics[width=\linewidth]{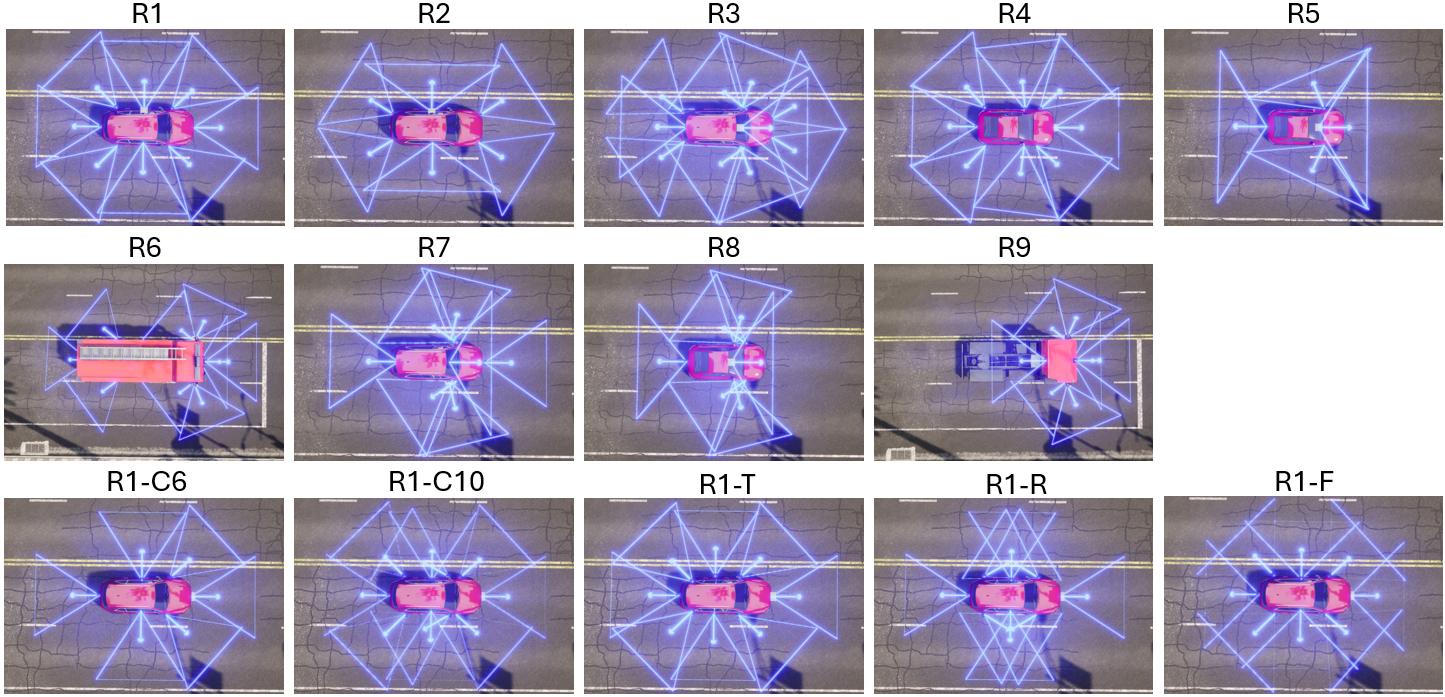}
  \caption{The rigs of our Plentiful Carla Camera Rigs benchmark. It contains nine unique rigs (top/center) and five factor-modified control rigs of R1 (bottom). Shown are the camera FOVs and the view directions.}
  \label{fig:rigs}
\end{figure}

To address this limitation, we introduce \emph{Plentiful Carla Camera Rigs}, a benchmark in which identical scenes are rendered under 14 systematically designed camera rigs, shown in Figure~\ref{fig:rigs}.
By keeping scene content, object layout, and environmental conditions fixed while varying only calibration parameters, the benchmark isolates rig-induced geometric effects.
Importantly, this setting does not require modeling camera-specific photometric characteristics or hardware artifacts.
Because our goal is to study \emph{relative performance shifts induced by geometric observation changes}, a high-fidelity simulator such as CARLA is sufficient: it allows controlled manipulation of extrinsics and intrinsics while holding all other factors constant.

This controlled benchmark enables systematic analysis of how camera rig geometry influences perception performance. 
To facilitate such analysis, we further introduce two rig-centric descriptors derived from calibration metadata.
First, \emph{Rig Variance} summarizes the internal geometric diversity and overlap structure of a single rig.
Second, \emph{Rig Contrastive Distance} measures geometric discrepancy between two rigs in terms of camera placement, orientation, FOV and coverage.
They provide a structured way to reason about how much the viewing geometry changes when transferring between rigs.
We empirically evaluate the relationship between these geometric descriptors and observed performance shifts across multiple camera-based 3D perception architectures.

Our contributions are summarized as follows:

\begin{itemize}
    \item We introduce \emph{Plentiful Carla Camera Rigs}, a controlled benchmark featuring identical scenes rendered under 14 systematically varied camera rigs.
    \item We propose two calibration-metadata-based geometric descriptors, Rig Variance and Rig Contrastive Distance, for quantifying rig properties and cross-rig discrepancies independent of any specific perception model.
    \item We provide empirical analysis demonstrating that geometric rig differences systematically relate to relative cross-rig performance shifts.
\end{itemize}

Any relevant supplementary information can be accessed through our project page at \url{https://badertim.github.io/plentiful-carla-camera-rigs/}.
\section{Related Work}
\label{sec:related_work}

We highlight existing work about rig descriptors, followed by rig-agnostic architectures and relevant data resources.

\subsection{Rig Configuration and Viewpoint Diversity}

Existing work on sensor placement for autonomous vehicles primarily evaluates configurations through task-dependent performance metrics such as coverage~\cite{indu2020optimal}, detection accuracy~\cite{wakabayashi2024camera}, or information gain~\cite{wang2004entropy}. 
A prominent line of research introduces \emph{perception entropy}~\cite{ma2024perceptionentropy, gamage2025perceptionframework}, which quantifies the expected uncertainty reduction provided by a given multi-sensor rig and has been applied to both camera--LiDAR configurations and optimization of roadside sensors. 
Closely related LiDAR-centric studies analyze non-detectable subspaces using geometric measures such as the volume-to-surface-area ratio (VSR)~\cite{liu2022lidarplacement}, or propose information-theoretic surrogates to predict 3D detection performance under different placements~\cite{hu2022multilidar,li2024cameralidar}. 
While these approaches demonstrate the strong impact of configuration on downstream perception, they are inherently tied to specific models, datasets, or environment assumptions.

A complementary body of work focuses on \emph{viewpoint quality} and \emph{view diversity}, most notably through viewpoint entropy~\cite{vazquez2003viewpointentropy, vazquez2003automatic, li2005information} to power several downstream tasks using optimized view selection. 
These methods capture how informative individual viewpoints are but do not provide a unified, rig-level descriptor for multi-camera systems. 

\subsection{Rig-Agnostic Perception Architectures}
\label{sec:related_work:rig_agnostic}
To overcome fixed-sensor constraints, recent work has focused on decoupling 2D feature extraction from 3D geometric reasoning.

\paragraph{Geometric Projection and Lifting.}
Lift-Splat-Shoot (LSS)~\cite{philion2020lift} introduced explicit view transformation by lifting image features into 3D frustums and projecting them onto a BEV grid via calibration, enabling geometric consistency across camera rigs. 
Building on this idea, \emph{BEVDet}~\cite{huang2021bevdet} improves robustness through independent image and BEV augmentations, while \emph{BEVFusion}~\cite{liang2022bevfusion} extends the approach to camera–LiDAR fusion. 
The \emph{Detecting As Labeling} (DAL)~\cite{huang2024detecting} framework further mitigates overfitting by separating candidate generation and regression across modalities.

\paragraph{Query-Based and Implicit Representations.}
Alternative approaches use sparse queries or implicit encodings to connect 2D observations and 3D reasoning. 
\emph{DETR3D}~\cite{wang2022detr3d} projects 3D object queries onto image features, while \emph{PETR}~\cite{liu2022petr, liu2023petrv2} injects 3D positional embeddings into image features to enable implicit geometric reasoning. 
Architectures such as \emph{BEVFormer}~\cite{li2024bevformer, yang2023bevformer} instead rely on learned spatial cross-attention, which can be more sensitive to rig changes.

\paragraph{Unconstrained Multi-View Reconstruction.}
Another direction explores unconstrained multi-view reconstruction from arbitrary image collections. 
Methods such as DUSt3R~\cite{wang2024dust3r}, VGGT~\cite{wang2025vggt}, and Fast3R~\cite{yang2025fast3r} demonstrate strong geometric reasoning under highly variable camera layouts. 
However, these approaches focus on scene reconstruction rather than semantic perception. 
Automotive-specific work such as Rig3R~\cite{li2025rig3r} conditions reconstruction on rig information but remains closed-source and difficult to evaluate reproducibly.

\subsection{Data and Benchmarks}
\label{sec:related_work:data}
Progress in automotive perception has been closely tied to large datasets and benchmarks. 
Most 3D object detection methods~\cite{mao20233d} are evaluated on datasets such as nuScenes~\cite{caesar2020nuscenes} and the Waymo Open Dataset~\cite{sun2020scalability}, which provide diverse driving scenarios together with fixed sensor configurations. 
While these benchmarks are essential for comparing perception architectures using metrics such as mean Average Precision (mAP) and the nuScenes Detection Score (NDS), their static sensor rigs prevent controlled analysis of how changes in camera placement, orientation, FOV, or camera count affect perception performance.

Simulation environments provide a promising alternative, as they allow explicit control over sensor calibration and scene generation. 
The CARLA simulator~\cite{Dosovitskiy17} enables rendering identical scenes under different sensor configurations while keeping all other factors constant. 
Using this capability, Embacher et al.~\cite{embacher_neural_2025} introduced a dataset in which an initial rig was fitted to two different vehicle types, demonstrating the existence of a cross-rig domain gap.
Related work has benchmarked viewpoint robustness by perturbing the cameras of an initial rig according to several viewpoint factors~\cite{klinghoffer_bevviewpointrobustness_2023}.
However, existing efforts remain limited in scale and do not provide a systematic benchmark covering a broader spectrum rig variations.

Recent advances in simulation and synthetic data generation further expand the potential for controlled perception benchmarks~\cite{paulin2023review, dalal2024gaussian}. 
Frameworks such as NVIDIA Cosmos~\cite{nvidia2025worldsimulationvideofoundation} and AlpaSim~\cite{alpasim_2025} enable scalable synthetic data generation but typically focus on increasing scene diversity rather than systematically varying sensor configurations.

Overall, existing datasets and simulation platforms either rely on fixed sensor rigs or provide unstructured synthetic diversity, limiting the study of rig-induced performance variation and motivating controlled multi-rig benchmarks.
\section{Plentiful CARLA Camera Rigs Benchmark}
\label{sec:benchmark}

The goal of the Plentiful CARLA Camera Rigs benchmark is to enable controlled analysis of cross-rig generalization in automotive perception. 
To isolate the effect of camera rig geometry, we render identical driving scenes under multiple systematically designed sensor configurations while keeping all other factors constant. 
This setup allows direct comparison of perception performance across rigs and enables reproducible study of how changes in camera placement, orientation, FOV, and camera count influence downstream models.

\subsection{Data Generation Process}
The dataset is generated with the CARLA Simulator~\cite{Dosovitskiy17} 0.9.16 (+ map expansion) using a two-stage simulation pipeline to ensure reproducibility and controlled variation across camera rigs. 
We first define a global metadata configuration specifying vehicle and sensor rigs, maps, weather conditions, dataset splits, and simulation frame rate (shown in Table~\ref{tab:dataset_summary} \& \ref{tab:sensor_rigs}).
Based on this configuration, we randomly sample scene descriptors for the training, validation, test and mini splits. 
Each scene descriptor specifies map, weather condition, traffic density, and ego-vehicle spawn point. 
Splits are defined at the scene level to prevent data leakage.

For each scene descriptor, the simulator is initialized and populated with dynamic traffic using the traffic manager, while the ego vehicle operates in autopilot mode. 
As the default traffic manager is non-deterministic, we record full trajectories of all simulated entities in an initial pass. 
Trajectory recording is performed using the largest vehicle model used in the dataset to ensure compatibility across sensor configurations with different vehicles.

Before replay, we apply a trajectory-level pruning step to reduce the over-representation of near-stationary ego-vehicle behavior by computing per-scene mean ego speed, binning scenes by speed, and downsampling over-populated bins.
The initial pool contains 100 trainval scenes and 40 test scenes, and pruning is configured to keep 80 trainval and 30 test scenes.

Next, all the kept scenes are replayed deterministically using the recorded trajectories. 
Each vehicle and sensor rig replays the same set of scenes, and the corresponding sensor data is recorded. 
This guarantees that different camera rigs observe identical scene dynamics.

Finally, all sensor data, 3D object annotations, and ego-vehicle metadata are stored in a standardized format inspired by nuScenes~\cite{caesar2020nuscenes}. 
Dependent on the target format, the data is processed further to fit MMDetection3D~\cite{mmdet3d2020}.

\subsection{Benchmark Statistics}
\label{sec:benchmark_stats}

The benchmark contains 115 scenes in total, where each has a fixed duration of 30\,s and is recorded at 2\,Hz, resulting in 60 samples per scene.
Across all splits, this yields 6{,}900 samples.
Considering the 105 cameras over all 14 rigs shown in Table~\ref{tab:sensor_rigs} and Figure~\ref{fig:rigs}, this leads to 724{,}500 images in total.

The dataset provides 3D annotations for nine object categories: car, truck, bus, motorcycle, bicycle, adult, child, traffic sign, and traffic light.
In the case of R1-c10, the benchmark contains 179{,}926 annotated objects corresponding to 7{,}299 unique instances.
Annotations include oriented 3D bounding boxes, semantic attributes, and discrete visibility labels that distinguish between 99{,}308 fully visible, 18{,}316 mostly visible and 42{,}283 partially visible objects dependent on the pixel coverage of the cameras.
This leads to minor annotation differences between rigs (e.g., R1-c6 has 159{,}907 annotations) to prevent ill-posed labels.
The annotations are limited up to a range of 80\,m from the ego-vehicle origin and all camera images are rendered at a fixed resolution of $1280\times720$ pixels.

The ego vehicle operates in autopilot mode and exhibits a diverse speed profile across scenes.
Over the full benchmark, ego-vehicle speeds range from 0 to 47.99\,km/h, with a mean speed of 16.02\,km/h and significant distribution peaks at 0\,km/h and 30\,km/h.
Table~\ref{tab:dataset_summary} shows more detailed information.

\begin{figure}[t]
  \centering
  \includegraphics[width=\linewidth]{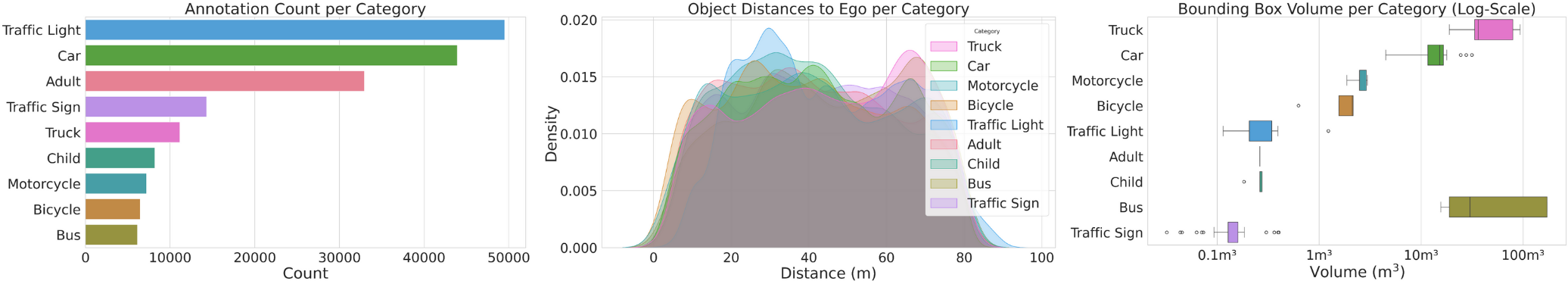}
  \caption{Benchmark distribution of all accumulated splits of R1-c10. Shown are the category distribution (left), the object distance distribution to the ego vehicle (center) and the 3D bounding box volume distribution (right). The distribution clusters of the latter can be attributed to the different category types.}
  \label{fig:distributions}
\end{figure}

\begin{table}[t]
\centering
\small
\begin{tabular}{|l|c|c|c|c|c|c|}
\hline
\textbf{Split} &
\textbf{Scenes} &
\textbf{Samples} &
\textbf{Annotations} &
\textbf{Maps} &
\textbf{Weathers} &
\textbf{Traffic Density} \\
\hline
Train+Val & 80 & 4{,}800 & 124{,}402 & 5 & 13 & [0.2, 0.8] \\
Test      & 30  & 1{,}800 & 47{,}843 & 3 & 12 & [0.1, 0.9] \\
Mini      & 5   & 300     & 7{,}681  & 2 & 2  & [0.3, 0.7] \\
\hline
Combined       & 115 & 6{,}900 & 179{,}926 & 8 & 13 & [0.1, 0.9] \\
\hline
\end{tabular}
\caption{Dataset statistics summarizing scene count, sample count, object annotations, environmental diversity, and traffic density coverage over data splits. Annotation count slightly varies between rigs due to camera placement, shown here is R1-c10.}
\label{tab:dataset_summary}
\end{table}

\begin{table*}[t]
\centering
\scriptsize
\begin{tabular}{
    c c
    >{\raggedright\arraybackslash}p{4cm}
    >{\raggedright\arraybackslash}p{2cm}
    >{\raggedright\arraybackslash}p{3.7cm}}
\toprule
\textbf{ID} & \textbf{Vehicle} & \textbf{Mounting Positions} & \textbf{Cameras} & \textbf{Description} \\ 
\midrule
\multicolumn{5}{l}{\textit{Controlled Rigs}} \\ 
\cmidrule{1-2}
R1      & SUV & Bumpers, fenders, B-pillars & 8$\times$65° & Base configuration \\ 
R1-r    & SUV & Bumpers, fenders, B-pillars & 8$\times$65° & Base with rotation changes \\ 
R1-t    & SUV & Bumpers, fenders, B-pillars & 8$\times$65° & Base with translation changes \\ 
R1-f    & SUV & Bumpers, fenders, B-pillars & 8$\times$90° & Base with FOV changes \\ 
R1-c6    & SUV & Front bumpers, fenders, B-pillars & 6$\times$65° & Base with fewer cameras \\ 
R1-c10    & SUV & Front bumpers, fenders, B-pillars & 10$\times$65° & Base with more cameras \\ 
\midrule
\multicolumn{5}{l}{\textit{Diverse Rigs}} \\ 
\cmidrule{1-2}
R2   & SUV        & Bumpers, B-pillars & 4$\times$80°, 2$\times$110° & Large side-camera FOV \\ 
R3   & SUV        & Bumpers, fenders, windshields, mirrors & 10$\times$75° & High-density coverage with strong overlap \\ 
R4   & Sports     & Bumpers, fenders, mirrors & 8$\times$65° & Sports-car geometry with different mounting semantics \\ 
R5   & Sports     & Mirrors, windshields & 4$\times$120° & Few wide-angle cameras \\
R6   & Firetruck  & Bumpers, fenders, mirrors, front roof & 8$\times$75°, 1$\times$120° & Large vehicle extrema with strong frontal overlap\\ 
R7  & SUV     & Bumpers, front windshield & 6$\times$70°, 1$\times$120° & Strong frontal overlap \\ 
R8  & Sports  & Bumpers, mirrors, front windshield & 4$\times$65°, 1$\times$100°, 1$\times$120° & Mixed front/back wide FOV \\ 
R9  & EU HGV  & Bumpers, front/back roof, mirrors, fenders & 6$\times$75°, 1$\times$120° & Tall-vehicle extrema with strong frontal overlap and roof-mounted front/back cameras \\ 

\bottomrule
\end{tabular}
\caption{
Sensor rig configurations used in our benchmark.  
The \emph{Controlled SUV Family} (R1, R1-r, R1-t, R1-f, R1-c) varies exactly one factor at a time: rotation, translation, FOV, or camera count. This enables a clean validation of our rig-difference metrics.  
The \emph{Diverse Rigs} (R2–R9) provide realistic variation in camera count, FOV mix, mounting semantics, and vehicle geometry for training and validation.  
}
\label{tab:sensor_rigs}
\end{table*}
\section{Rig Variation Metrics}
\label{sec:metrics}

Camera rigs for autonomous driving vary in placement, orientation, and FOV, and such variations strongly influence model performance. 
We introduce two metrics that quantify these differences based on camera poses and intrinsics. 

The metrics are complementary: Rig Variance characterizes the internal heterogeneity of a single rig, while Rig Contrastive Distance quantifies the discrepancy between two distinct rigs. 
Together, they enable a preliminary analysis of how changes such as translation shifts, rotational perturbations, or FOV modifications affect perception models.

\subsection{Rig Variance}

Rig Variance (RigV) measures the internal heterogeneity of a single rig and captures how diverse the cameras are in their spatial and optical configuration. 
Each camera $i$ is represented as $\mathbf{c}_i = (\mathbf{t}_i, \mathbf{r}_i, f_i)$, where $\mathbf{t}_i \in \mathbb{R}^3$ denotes the position in ego coordinates, $\mathbf{r}_i \in SO(3)$ denotes the orientation, and $f_i$ denotes the FOV 
RigV computes the average pairwise divergence within a rig
\begin{equation}
\text{RigV} = \frac{1}{N(N-1)} \sum_{i \neq j} D(\mathbf{c}_i, \mathbf{c}_j),
\end{equation}
with
\begin{equation}
D(\mathbf{c}_i, \mathbf{c}_j) =
\lambda_t \, \Delta t_{ij}
+ \lambda_r \, \Delta r_{ij}
+ \lambda_f \, \Delta f_{ij}.
\end{equation}
Here, the individual components are defined as
\begin{equation}
\Delta t_{ij} = \frac{\|\mathbf{t}_i - \mathbf{t}_j\|_2}{D_{\max}}, \quad
\Delta r_{ij} = \frac{d_R(\mathbf{r}_i, \mathbf{r}_j)}{\pi}, \quad
\Delta f_{ij} = \frac{|f_i - f_j|}{180}
\end{equation}
where $d_R$ denotes the geodesic rotation distance on $SO(3)$. The translation normalization factor
\begin{equation}
D_{\max} = \max_{k,l} \|\mathbf{t}_k - \mathbf{t}_l\|_2
\end{equation}
corresponds to the maximum pairwise camera distance within the same rig.

This normalization ensures that all components $\Delta t_{ij}$, $\Delta r_{ij}$, and $\Delta f_{ij}$ lie in the range $[0,1]$, making the metric scale-consistent across translation, rotation, and field-of-view differences. 
The weighting parameters $\lambda_t, \lambda_r, \lambda_f$ control the relative contribution of the normalized translation, rotation, and FOV terms.

RigV increases with positional spread, rotational disparity, and FOV heterogeneity, yielding a compact descriptor of geometric rig diversity. 
Higher RigV values indicate rigs with stronger internal variation, which may provide broader scene coverage while simultaneously increasing perception difficulty.

\subsection{Rig Contrastive Distance}
Rig Contrastive Distance (RigCD) measures the geometric and optical disparity between two rigs $A$ and $B$. Unlike RigV, which characterizes the internal heterogeneity of a single configuration, RigCD directly quantifies cross-rig discrepancy and reflects how much a model trained on one rig must adapt to another.

Each camera is represented by its 3D position, full 3D orientation, and FOV. For every camera pair across rigs, we construct the cost matrix
\[
C_{ij}
=
\lambda_t \|\mathbf{t}_i^A-\mathbf{t}_j^B\|_2
+ \lambda_r d_R(\mathbf{r}_i^A,\mathbf{r}_j^B)
+ \lambda_f |f_i^A-f_j^B|,
\]
where $d_R$ is the geodesic rotation distance on $SO(3)$. Unlike RigV, this cross-rig matching cost uses the raw translation, rotation, and FOV differences directly.
To resolve camera correspondences, we compute the optimal bipartite assignment over the two rigs via minimum-cost matching. 
Let $N_A$ and $N_B$ denote the number of cameras in rigs $A$ and $B$, and let $M = \min(N_A, N_B)$. 
The matched discrepancy is

\begin{equation}
\text{RigCD}_{\text{match}}(A, B)
=
\frac{1}{M} \sum_{(i,j) \in \mathcal{M}}
C_{ij},
\end{equation}

where $\mathcal{M}$ is the optimal set of $M$ matched camera pairs. 
Since rigs may contain different camera counts, we introduce a normalized count-mismatch penalty

\begin{equation}
\text{RigCD}_{\text{count}}(A, B)
=
\frac{|N_A - N_B|}{\max(N_A, N_B)},
\end{equation}

which increases when one rig provides substantially more or fewer viewpoints than the other. The final Rig Contrastive Distance combines these terms:

\begin{equation}
\text{RigCD}(A, B) = \alpha\, \text{RigCD}_{\text{match}}(A, B) + (1 - \alpha)\, \text{RigCD}_{\text{count}}(A, B),
\end{equation}

with $\alpha \in [0,1]$ controlling the relative importance of geometric alignment vs. camera-count discrepancy.
Componentwise variants $\text{RigCD}_t$, $\text{RigCD}_r$, $\text{RigCD}_f$ and $\text{RigCD}_{\text{count}}$ isolate differences in translation, rotation, FOV and count, enabling fine-grained attribution of performance degradation under rig shifts.

\section{Evaluation Protocol}
\label{sec:experiments}
In this section we describe the evaluation protocol used in the Plentiful CARLA Camera Rigs benchmark. 
The protocol measures cross-rig generalization by training perception models on one camera configuration and evaluating them on others. 
We further outline the baseline models used for evaluation and the calibration procedure for the proposed rig-distance metrics.

\subsection{Cross-Rig Evaluation Protocol}
The benchmark evaluates how perception models generalize across camera rigs.
Given a set of sensor rigs $\mathcal{R}=\{R1,\dots,R9\}$, a model is trained on the training split of a source rig $R_i$ and evaluated on the test split of a target rig $R_j$.

This yields a cross-rig transfer matrix $mAP_{\mathrm{rel}}(R_i \rightarrow R_j)$ over all source–target rig combinations, from which we compute the relative generalization gap
\begin{equation}
\Delta mAP_{\mathrm{rel}}(R_i \rightarrow R_j) 
= \frac{mAP(R_i \rightarrow R_i) - mAP(R_i \rightarrow R_j)}
       {mAP(R_i \rightarrow R_i)}.
\end{equation}
This matrix represents the empirical performance degradation caused by changes in sensor stack configuration.

\subsection{Baseline Models}
To provide reference performance on the benchmark, we evaluate four representative camera-based multi-view 3D detection architectures: BEVDet~\cite{huang2021bevdet}, BEVFusion~\cite{liang2022bevfusion}, Fast-BEV~\cite{li2024fast}, and PETR~\cite{liu2022petr}.
BEVDet and BEVFusion lift image features into an explicit BEV representation using LSS-style view transformation, Fast-BEV projects multi-view image features into a 3D voxel representation for efficient BEV perception, and PETR performs DETR-style object detection by enriching multi-view image features with 3D position embeddings.

All models are trained using their officially released MMDetection3D-based codebases~\cite{mmdet3d2020}.
We adapt the data loaders and nuScenes-style train/evaluation pipelines to the PCCR benchmark format while keeping the original model configurations as close as possible.
All experiments use camera input only and predict the nine PCCR benchmark classes.
Velocity targets are disabled where supported by the baseline configuration, and velocity is discarded during evaluation.
Training is performed with AdamW~\cite{loshchilov2017decoupled} on two NVIDIA RTX~6000 GPUs with a batch size of four per GPU.

\subsubsection{BEVDet}
We use the camera-only BEVDet configuration with a ResNet-50~\cite{he2016deep} image backbone, a CustomFPN image neck, an LSS view transformer, a BEV encoder, and a CenterPoint-style CenterHead detection head~\cite{yin2021center}.
Images are resized to $384\times704$, and the model is trained for 24 epochs.

\subsubsection{BEVFusion}
For BEVFusion, we use the camera-only configuration.
The LiDAR encoder and fusion module are disabled, so the model consists of a ResNet-50 image encoder, an SECONDFPN image neck, an LSS-based view transformation module, a BEV decoder, and a CenterHead detector.
Images are resized to $256\times704$, and the model is trained for 20 epochs.

\subsubsection{Fast-BEV}
For Fast-BEV, we use the multi-frame camera-only configuration with a ResNet-50 backbone, FPN neck, voxel-based multi-view transformation, an M2BEV neck, and an anchor-based FreeAnchor3D detection head.
Images are resized to $320\times576$, and the model is trained for 40 epochs.

\subsubsection{PETR}
For PETR, we use the camera-only PETR3D configuration with GridMask augmentation, a ResNet-50-DCN image backbone, a CPFPN neck, and a PETR transformer decoder.
The PETR head uses 900 object queries, 3D sine positional encoding, and multi-view transformer cross-attention.
Images are resized to $320\times800$, and the model is trained for 120 epochs.

\subsection{Rig Metric Calibration}
\label{sec:experiments:rig_metric_calib}
The weighting parameters of RigV and RigCD are calibrated using the controlled stacks derived from $R1$.
For each variant combination, we use the previously measured $\Delta mAP_{\mathrm{rel}}(R_i \rightarrow R_j)$
along with the component-wise distances 
$(\text{RigCD}_t,\text{RigCD}_r,\text{RigCD}_f,\text{RigCD}_{\text{count}})$.
We apply L-BFGS-B optimization and obtain the calibrated parameters 
$\boldsymbol{\lambda}=\{\lambda_t,\lambda_r,\lambda_f\}$ and $\alpha$, 
together with a global scale factor $K$, yielding the calibrated prediction
\begin{equation}
\widehat{\Delta mAP}_{\mathrm{rel}}(R_i \rightarrow R_j)
=
K \cdot \text{RigCD}(R_i,R_j;\alpha,\boldsymbol{\lambda}).
\end{equation}
To quantify how well the calibrated metric preserves the ordering of performance degradations under isolated perturbations, we report Spearman rank correlation $\rho$ between $\widehat{\Delta mAP}_{\mathrm{rel}}$ and the observed $\Delta mAP_{\mathrm{rel}}$, and compute a $95\%$ confidence interval via bootstrap resampling (sampling stacks with replacement, $5{,}000$ iterations).
We evaluate the ranking on the held-out sensor rigs $\{R2,\dots,R8\}$, which are not used during calibration.

\section{Benchmark Results}
\label{sec:results}
We present the main findings obtained using the PCCR benchmark. 
Following the evaluation protocol, we analyze cross-rig transfer behavior and examine how geometric rig differences influence perception performance.

\subsection{Cross-Rig Generalization Analysis}
Following the cross-rig evaluation protocol described in Section~\ref{sec:experiments}, we first report in-domain performance, where each model is trained and evaluated on data from the same rig.
Averaged across all rigs, BEVFusion obtains the highest mAP with $0.163 \pm 0.052$, followed by BEVDet with $0.155 \pm 0.053$, PETR with $0.149 \pm 0.049$, and Fast-BEV with $0.090 \pm 0.043$.
These results show that the proposed benchmark supports training and evaluating diverse camera-based 3D detection architectures, even when using compact baseline configurations.

\begin{figure}[t]
  \centering
  \includegraphics[width=\linewidth]{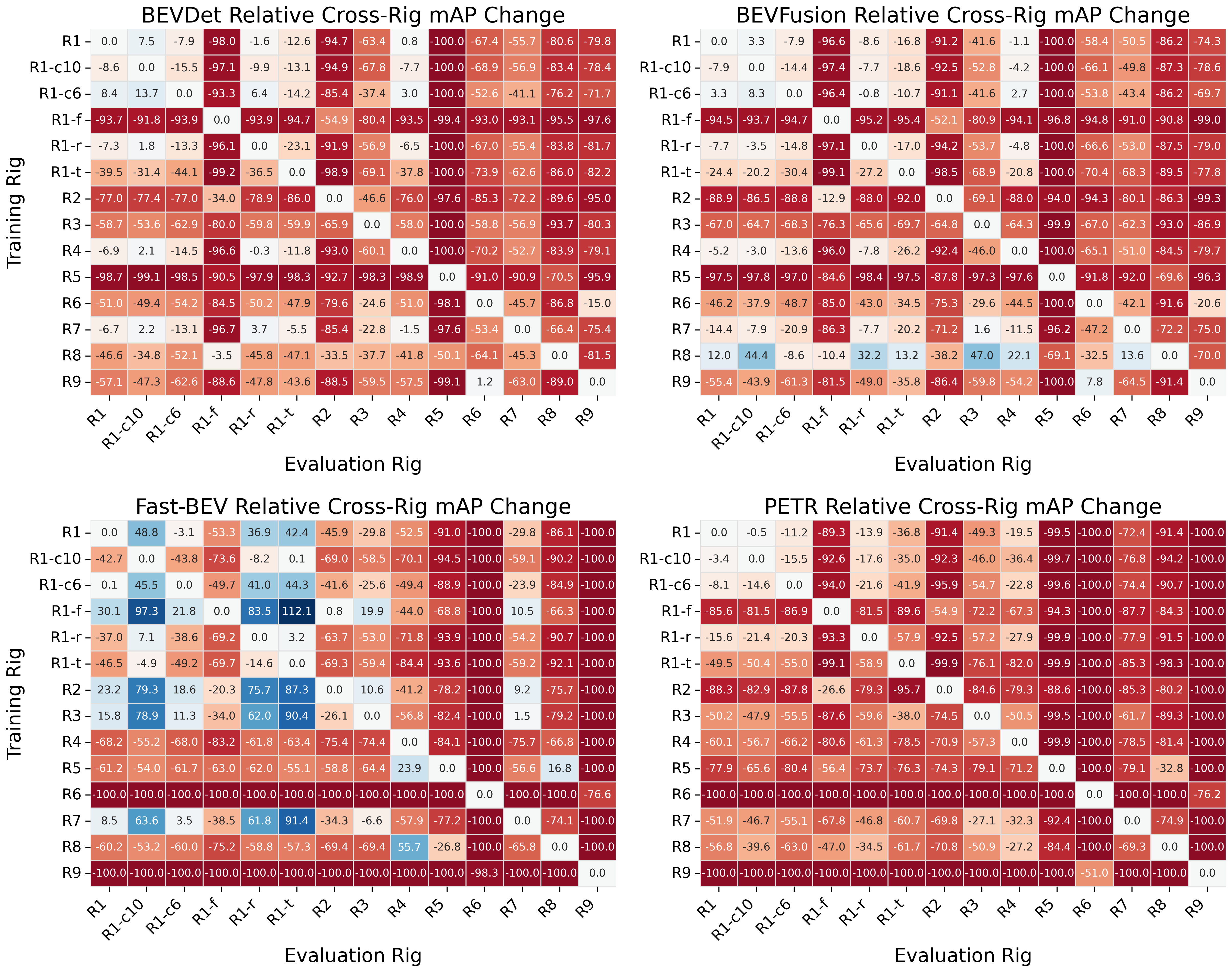}
  \caption{Relative generalization gaps between rigs, based on mAP. Y-axis shows the rig trained on and X-axis shows the evaluated rigs.}
  \label{fig:relative_map_change}
\end{figure}

Figure~\ref{fig:relative_map_change} visualizes the resulting cross-rig transfer matrices
$\Delta mAP_{\mathrm{rel}}(R_i \rightarrow R_j)$.
These results demonstrate that cross-rig domain gaps can be substantial even when the scene content remains identical.
Under out-of-distribution (OOD) FOV changes, BEVDet, BEVFusion and PETR experience a near total collapse.
We observe similar performance drops using the more complex rigs R6 and R9 with Fast-BEV and PETR.
However, training on these higher-variance rigs (R6--R9) improves cross-rig generalization for BEVFusion and BEVDet, leading to smaller relative gaps on unseen rigs.
Overall, we observe that cross-rig robustness is highly architecture-dependent.

\subsection{Metric Calibration and Evaluation}

\begin{table*}[t]
\scriptsize
\centering
\resizebox{\linewidth}{!}{
\begin{tabular}{lccccc|cc|cc}
\toprule
& \multicolumn{5}{c|}{Calibration parameters}
& \multicolumn{2}{c|}{Controlled Rigs}
& \multicolumn{2}{c}{Diverse  R igs} \\
\cmidrule(lr){2-6} \cmidrule(lr){7-8} \cmidrule(lr){9-10}
Model
& $\alpha$ & $K$ & $\lambda_t$ & $\lambda_r$ & $\lambda_f$
& $\rho$ & CI
& $\rho$ & CI \\
\midrule
BEVDet
& 0.679 & 0.232 & 2.427 & 0.346 & 0.227
& 0.866 & [0.726, 0.926]
& 0.718 & [0.620, 0.791] \\

BEVFusion
& 0.494 & 0.257 & 2.178 & 0.537 & 0.285
& 0.914 & [0.824, 0.949]
& 0.734 & [0.649, 0.800] \\

Fast-BEV
& 0.900 & 1.000 & 1.000 & 1.000 & 1.000
& 0.052 & [-0.339, 0.423]
& 0.097 & [-0.054, 0.235] \\

PETR
& 0.409 & 0.685 & 2.284 & 0.609 & 0.108
& 0.982 & [0.946, 0.993]
& 0.804 & [0.735, 0.851] \\
\bottomrule
\end{tabular}
}
\caption{RigCD calibration parameters and rank-correlation performance on the calibration (controlled rigs) and test (diverse rigs) splits. We report Spearman's $\rho$ with 95\% confidence intervals (CI).}
\label{tab:calibration_results}
\end{table*}

Using the controlled perturbation rigs, we calibrate RigCD and analyze how geometric factors affect cross-rig transfer in Table~\ref{tab:calibration_results}.
Across the reliable baselines, the learned parameters consistently assign the largest weight to camera translation changes, with $\lambda_t=2.43$ for BEVDet, $\lambda_t=2.18$ for BEVFusion, and $\lambda_t=2.28$ for PETR.
Rotation receives a moderate weight for BEVFusion and PETR ($\lambda_r=0.54$ and $0.61$), while FOV changes are weighted lower, especially for PETR ($\lambda_f=0.11$).

RigCD achieves strong ranking agreement on the calibration set for BEVDet, BEVFusion, and PETR, with $\rho=0.87$, $0.91$, and $0.98$, respectively.
On unseen rig pairs, the correlations remain strong: $\rho=0.72$ for BEVDet, $\rho=0.73$ for BEVFusion, and $\rho=0.80$ for PETR, with 95\% CIs of $[0.62,0.79]$, $[0.65,0.80]$, and $[0.73,0.85]$.
Fast-BEV is an outlier, with weak calibration and test correlations ($\rho=0.05$ and $0.10$), suggesting that RigCD does not reliably explain its cross-rig behavior under this configuration.

\begin{figure}[t]
  \centering
  \includegraphics[width=\linewidth]{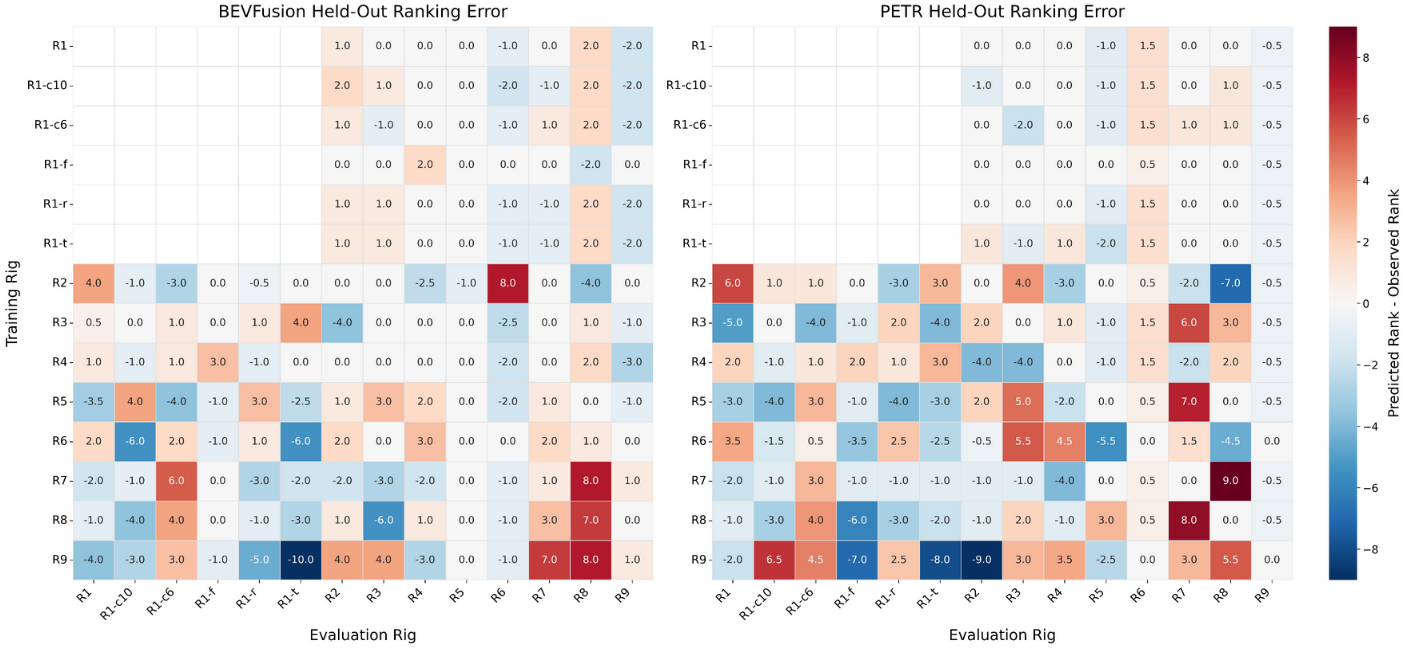}
  \caption{Signed ranking error heatmap for RigCD-based transfer prediction. Each cell shows the difference between predicted and observed target-rig rank within the test rig set, computed as predicted rank minus observed rank. Values near zero indicate better agreement. Positive and negative values show the direction of rank mismatch.}
  \label{fig:calibrated_rigcd_ranking_error}
\end{figure}

Overall, the calibrated RigCD metric indicates that cross-rig transfer is primarily driven by camera translation differences, followed by rotation, while FOV differences receive lower effective weights for the best-correlated models.
While the absolute prediction of mAP remains a challenging and somewhat unstable task due to environmental factors, the consistent ranking performance shown in Figure~\ref{fig:calibrated_rigcd_ranking_error} demonstrates that RigCD provides a dependable proxy for relative rig difficulty.

We show the calibration impact on RigV in Table~\ref{tab:rigv_variants} and observe that cross-rig evaluations on rigs with higher RigV correlate with decreasing mAP. 

\begin{table*}[t]
\centering
\scriptsize
\setlength{\tabcolsep}{3pt}
\begin{tabular}{lcccccc|cccccccc}
\toprule
Calibration
& \multicolumn{6}{c|}{Controlled Rigs (R1)}
& \multicolumn{8}{c}{Diverse Rigs} \\
\cmidrule(lr){2-7} \cmidrule(lr){8-15}
& R1 & C10 & C6 & F & R & T
& R2 & R3 & R4 & R5 & R6 & R7 & R8 & R9 \\
\midrule
None
& 1.09 & 1.02 & 1.08 & 1.09 & 1.09 & 1.07
& 1.30 & 1.06 & 1.14 & 1.33 & 1.13 & 1.09 & 1.29 & 1.37 \\

BEVDet
& 1.46 & 1.36 & 1.44 & 1.46 & 1.46 & 1.40
& 1.72 & 1.44 & 1.59 & 1.94 & 1.50 & 1.28 & 1.62 & 1.98 \\

BEVFusion
& 1.44 & 1.34 & 1.42 & 1.44 & 1.44 & 1.38
& 1.69 & 1.42 & 1.55 & 1.88 & 1.47 & 1.29 & 1.60 & 1.92 \\

Fast-BEV
& 1.09 & 1.02 & 1.08 & 1.09 & 1.09 & 1.07
& 1.30 & 1.06 & 1.14 & 1.33 & 1.13 & 1.09 & 1.29 & 1.37 \\

PETR
& 1.53 & 1.43 & 1.52 & 1.53 & 1.53 & 1.48
& 1.78 & 1.51 & 1.66 & 2.00 & 1.56 & 1.36 & 1.68 & 2.02 \\
\bottomrule
\end{tabular}
\caption{Rig variance (RigV) values across rigs under different calibration variants.}
\label{tab:rigv_variants}
\end{table*}

\subsection{Ablation Studies}

We conduct ablations to further analyze our benchmark and the metrics.

\paragraph{Component contribution analysis.}
We perform a component knockout study by vanishing each calibrated weight $(\lambda_t, \lambda_r, \lambda_f, \text{and } \alpha)$ in turn and re-evaluating the predictive performance on unseen rigs; the results are shown in Table~\ref{tab:component_ablation}.
Across BEVDet and BEVFusion, the RigCD metric relies most strongly on FOV/overlap consistency: removing $\lambda_f$ reduces Spearman correlation by $0.39$ and $0.40$, respectively.
PETR remains most sensitive to translation errors, with $\lambda_t=0$ reducing $\rho$ by $0.58$.
Rotation and camera-count terms have comparatively small effects for most models, suggesting that their calibrated contributions are secondary once translation and overlap structure are accounted for.
Fast-BEV exhibits a low full-model correlation and improves when the FOV/overlap term is removed, indicating that its observed degradation pattern is not well explained by the same FOV/overlap weighting that benefits the other BEV-based models.
Overall, these findings show that different architectures emphasize different geometric factors, while the integrated RigCD formulation captures the dominant sources of rig-induced domain shift for most models.

\begin{table}[t]
\centering
\scriptsize
\setlength{\tabcolsep}{3pt}
\begin{tabular}{lcc|cc|cc|cc}
\toprule
\multirow{2}{*}{RigCD Metric Variant} &
\multicolumn{2}{c}{BEVDet} &
\multicolumn{2}{c}{BEVFusion} &
\multicolumn{2}{c}{Fast-BEV} &
\multicolumn{2}{c}{PETR} \\
& $\rho$ & $\Delta\rho$ & $\rho$ & $\Delta\rho$ & $\rho$ & $\Delta\rho$ & $\rho$ & $\Delta\rho$ \\
\midrule
Full Model & 0.718 & 0.000 & 0.734 & 0.000 & 0.097 & 0.000 & 0.804 & 0.000 \\
w/o Translation ($\lambda_t=0$) & 0.743 & 0.025 & 0.719 & -0.015 & 0.013 & -0.084 & 0.220 & -0.584 \\
w/o Rotation ($\lambda_r=0$) & 0.719 & 0.001 & 0.738 & 0.004 & 0.092 & -0.005 & 0.808 & 0.003 \\
w/o FOV/Overlap ($\lambda_f=0$) & 0.330 & -0.388 & 0.331 & -0.403 & 0.679 & 0.582 & 0.787 & -0.018 \\
w/o Camera Count ($\alpha=1$) & 0.716 & -0.002 & 0.733 & -0.002 & 0.097 & 0.000 & 0.800 & -0.005 \\
\bottomrule
\end{tabular}
\caption{Component ablation for RigCD. We remove individual metric components and measure the resulting Spearman correlation ($\rho$) between predicted rig distance and observed relative performance degradation. $\Delta\rho$ denotes the difference.}
\label{tab:component_ablation}
\end{table}

\paragraph{Multi-model metric calibration.}
To evaluate the model-agnostic applicability of RigCD, we calibrate and evaluate RigCD as in Section~\ref{sec:experiments:rig_metric_calib}, but process BEVDet, BEVFusion, Fast-BEV, and PETR jointly instead of separately.

The jointly calibrated RigCD achieves strong generalization on unseen rigs for BEVDet with $\rho=0.71$ (95\% CI: $[0.61, 0.78]$), BEVFusion with $\rho=0.68$ (95\% CI: $[0.58, 0.75]$), Fast-BEV with $\rho=0.59$ (95\% CI: $[0.48, 0.68]$), and PETR with $\rho=0.74$ (95\% CI: $[0.65, 0.80]$).
These results indicate that a shared RigCD calibration captures geometric rig differences across architectures, while the remaining variation reflects model-specific sensitivities to particular rig perturbations.
Interestingly, this has also significantly improved the results for Fast-BEV, suggesting that models that are harder to calibrate benefit from more data points, even when these are from different models.

\paragraph{Multi-rig training ablation.}
We study multi-rig training using BEVFusion on a non-overlapping union of rig slices ($R1$, $R2$, $R3$, $R4$, $R5$, $R6$), resulting in a dataset equivalent in size to a single rig.
Evaluating on unseen rigs ($R7$, $R8$, $R9$), we measure $\Delta mAP_{\mathrm{rel}}$ relative to the identity pairs (e.g., $R7 \rightarrow R7$). 
This yields $-38.37\%$ on $R7$, $-71.16\%$ on $R8$, and $-48.14\%$ on $R9$.
While this improves average cross-rig performance and variance, it does not surpass the best single-rig transfer cases.
\section{Discussion}
\label{sec:discussion}

\paragraph{General discussion.}
The benchmark shows that cross-rig generalization is highly architecture-dependent.
BEVFusion achieves the best in-domain performance, followed by BEVDet and PETR, while Fast-BEV performs noticeably worse overall.
Under cross-rig evaluation, BEVDet, BEVFusion, and PETR show substantial sensitivity to rig changes, with severe drops under out-of-distribution FOV shifts and on some complex diverse rigs.

RigCD provides a useful ranking proxy for most models.
After calibration, it correlates strongly with unseen cross-rig transfer difficulty for BEVDet, BEVFusion, and PETR, reaching $\rho=0.72$, $0.73$, and $0.80$, respectively.
Fast-BEV is an exception: its weak correlations indicate that its transfer behavior is not well explained by the same calibrated geometric factors.

The calibrated weights show that translation differences are the dominant factor for the reliable baselines, followed by rotation, while FOV receives lower effective weights.
The RigV results further indicate that higher geometric rig variance generally corresponds to more difficult transfer, supporting its use as a compact descriptor of rig complexity.

\paragraph{Limitations.}
The the benchmark intentionally isolates geometric rig variation by rendering identical scenes across rigs.
While this enables controlled analysis, it also implies that the benchmark remains simulation-based and therefore subject to the simulation-to-real gap.
Our baseline evaluation focuses on two representative  models, which limits the generality of architectural conclusions.

RigCD relies solely on calibration metadata and therefore captures only geometric aspects of the observation process.
Other sensor characteristics, such as photometric differences, lens distortion, or hardware-specific artifacts, are not modeled.
Furthermore, RigCD is defined for pairwise rig comparison and does not directly extend to scenarios where models are trained on multi-rig datasets.

\paragraph{Future work.}
Future work could extend the benchmark with additional sensor modalities, rig configurations, and environmental diversity.
Incorporating real-world datasets would further improve ecological validity.
From a modeling perspective, the results motivate approaches explicitly designed for cross-rig robustness, such as rig-aware data augmentation or calibration-conditioned perception models.
Another promising direction is uncertainty estimation when deploying perception systems on previously unseen sensor rigs.
\section{Conclusion}
\label{sec:conclusion}
We introduced the \textit{Plentiful CARLA Camera Rigs} benchmark for studying cross-rig generalization in camera-based automotive perception by rendering identical driving scenes under systematically varied camera configurations. 
Using this setup, we show that geometric rig variation can induce substantial performance shifts and that cross-rig robustness depends strongly on architectural design. 
We further propose two calibration-based descriptors, \textit{Rig Variance} and \textit{Rig Contrastive Distance} (RigCD), to quantify rig diversity and cross-rig discrepancy using calibration metadata. 
Our experiments show that RigCD correlates strongly with observed cross-rig performance degradation and provides a useful proxy for ranking transfer difficulty between sensor rigs.



%
%
\bibliographystyle{splncs04}
\bibliography{main}

\end{document}